\documentclass[a4paper]{easychair}
\usepackage{graphicx}
\usepackage{mathptmx}      
\usepackage{proof}
\usepackage{amssymb}
\usepackage{amsmath}
\usepackage{stmaryrd}
\usepackage{extarrows}
\usepackage{tikz}
\usepackage{array}
\usetikzlibrary{positioning}
\usepackage{hyperref}
\usepackage{caption}
\usetikzlibrary{matrix}
\usepackage{lscape}
\usepackage{booktabs}
\newcommand{\ra}[1]{\renewcommand{\arraystretch}{#1}}

\newcommand{\quotes}[1]{``#1''}
\newcommand{\singlequotes}[1]{`#1'}

\newcommand{\tensor}{\otimes}
\newcommand{\bs}{\backslash}
\newcommand{\s}{\slash}

\newcommand{\reals}{\mathbb{R}}

\newcommand{\KroneckerRho}[3]{\rho\kern-0.2em{\raisebox{0.6em}{ {\tiny $#1$} }} \kern-0.7em{\raisebox{0.15em}{ {\tiny $#2$} }} \kern-0.75em{\raisebox{-0.4em}{ {\tiny $#3$} }}}

\newcommand{\abs}[2]{\lambda #1. #2}

\newcommand{\vek}[1]{\overrightarrow{#1}}

\newtheorem{definition}{Definition}
\begin{document}

\title{A Typedriven Vector Semantics for Ellipsis with Anaphora using Lambek Calculus with Limited Contraction 
}


\author{Gijs Wijnholds\inst{1}         \and
        Mehrnoosh Sadrzadeh\inst{1} 
}


\institute{School of Electronic Engineering and Computer Science, \\
              Queen Mary University of London \\
              \email{g.j.wijnholds@qmul.ac.uk,m.sadrzadeh@qmul.ac.uk}           
}


\maketitle

\begin{abstract} 
We develop a vector space semantics for verb phrase ellipsis with anaphora using type-driven compositional distributional semantics based on the Lambek calculus with limited contraction (\textbf{LCC}) of J{\"a}ger \cite{jager2006anaphora}.  Distributional semantics has a lot to say about the statistical colocation-based  meanings of content words, but provides little guidance on how to treat function words. Formal semantics on the other hand, has powerful mechanisms for dealing with relative pronouns, coordinators, and the like. Type-driven compositional distributional semantics brings these two models together.  We review previous compositional distributional models of  relative pronouns, coordination and a restricted account of ellipsis in the DisCoCat framework of Coecke et al. \cite{coecke2010mathematical,coecke2013lambek}. We show how DisCoCat cannot deal with general forms of ellipsis, which rely on copying of information, and develop a novel way of connecting typelogical grammar to distributional semantics by assigning  vector interpretable   lambda terms to derivations of \textbf{LCC} in the style of Muskens \& Sadrzadeh \cite{muskens2016context}. What follows is an account of (verb phrase) ellipsis in which word meanings can be copied: the meaning of a sentence is now a program with non-linear access to individual word embeddings. We present the theoretical setting, workout examples, and demonstrate our results on a toy distributional model motivated by data. 
\end{abstract}

\newpage

\section{Introduction}
\label{intro}

Distributional semantics is a field of research within computational linguistics that provides an easily  implementable algorithm with an empirically verifiable output for representing word meanings and  degrees of semantic similarity thereof. This semantics is rooted in the distributional hypothesis,  often referred to via the quote \quotes{you shall know a word by the company it keeps},  made popular by Firth \cite{firth1957synopsis}. More precisely, according to the distributional hypothesis  words that occur in similar contexts have similar meaning. This idea has been made concrete by gathering the  co-occurrence statistics of context and target words  in  corpora of text and using that as a basis for developing vector representations for word meanings. A notion of similarity based on the cosine of the angle between vectors  allows one to compare degrees of word similarity in the vector space models where these word vectors embed. Such models have been shown to perform well in a variety of natural language processing (NLP) tasks, such as semantic priming \cite{lund1996producing} and word sense disambiguation \cite{schutze1998automatic}. The underlying philosophy has gained attention in cognitive science as well \cite{lenci2008distributional}.

Although this notion of similarity is intuitive and works well at the word level, it is less productive to consider phrases and full sentences to be similar whenever they occur in a similar context. Firstly, we know that language is compositional, since the number of potential sentences humans can produce are  larger than the amount a single human ever produces. Secondly, data sparsity issues arise when treating sentences as individual expressions and computing direct co-occurrence statistics for them. So the challenge of producing vector representations for phrases and sentences rests on the shoulders of compositional distributional semantics. Several studies have tried to learn not just vectors for words, but embeddings for several constituents \cite{baroni2010nouns,grefenstette2013multi}, or have experimented with simple commutative compositional operations such as addition and multiplication  \cite{mitchell2008vector}. A structured attempt at providing a general  mathematically sound model of compositional distributional semantics has been presented by Coecke et al. \cite{coecke2010mathematical}; these models start from the observation that vector spaces share the same structure as Lambek's most recent grammar formalism, pregroup grammar \cite{lambek1997type}, and interpret its derivations in terms of vector spaces and linear maps. What follows is an architecture that is familiar from logical formal semantics in Montague style \cite{montague1970english}, where the judgments of a grammar translate to a consistent semantic operation (read linear map) that acts on the individual word vectors to produce some vector in the sentence space. A number of subsequent  attempts has shown that a similar interpretation is possible for other typelogical grammars, such as Lambek's original syntactic calculus \cite{coecke2013lambek}, Lambek-Grishin grammars \cite{wijnholds2014categorical}, and the Combinatorial Categorial Grammar (CCG) \cite{maillard2014type}. 

One major issue for distributional semantics and especially compositional approaches therein  is to find a suitable representation for function words. Without the power of formal semantics to assign constant meanings or to allow set-theoretic operations, distributional semantics does not have much to say about the meaning of logical words such as \singlequotes{and}, \singlequotes{despite} and pronouns like \singlequotes{his}, \singlequotes{which}, \singlequotes{that}, let alone quantificational constituents (\singlequotes{all}, \singlequotes{some}, \singlequotes{more than half}). All of these words have in common that they intuitively do not bear a contextual meaning: a function word may co-occur with any content word and so its distribution does not reveal  much about  its meaning, unless perhaps the notion of meaning is taken to be conversational\footnote{The work of Kruszewski et al. \cite{kruszewski2016there} gives a distributional semantic account of conversational negation}. To overcome this issue, Sadrzadeh \cite{sadrzadeh2013frobenius} relies on Frobenius algebras to formalise the notion of \emph{combining} and \emph{dispatching} of information. This approach has seen applications to relative pronouns \cite{sadrzadeh2013frobenius,sadrzadeh2014frobenius}, coordination \cite{kartsaklis2016coordination}, and to a lesser extent to some limited forms of ellipsis \cite{kartsaklisverb}. In each of these,  the Frobenius algebras allow one to use element wise multiplication of arbitrary tensors, corresponding to the usual intersective interpretation one finds in formal semantics \cite{dalrymple1991ellipsis}. A treatment of quantification was also given using the bialgebraic nature of vector spaces over powersets of elements \cite{hedges2016generalised,sadrzadeh2016quantifier}. An explanation of the derivational processes resulting in these compositional meanings requires more elaborate grammatical mechanisms: Wijnholds \cite{wijnholds2014categorical} repeats the exercise to give a compositional distributional model for a symmetric extension of the Lambek calculus. A derivational account of pronoun relativisation in English and Dutch is given by means of a Lambek grammar with controlled forms of movement and permutation in \cite{moortgat2017lexical}.

In this paper, we contribute to the typelogical style of compositional distributional semantics by giving an account for verb phrase ellipsis with anaphora in a revision of the framework described above.
The case of ellipsis traditionally has been approached both as a syntactic problem within categorical grammars \cite{KubotaLevine2017,jager2006anaphora,jager1998multi,morrill1996generalising,Hendriks1995} as well as a semantic problem by directly  appealing to  their  lambda calculi  term logics \cite{dalrymple1991ellipsis,kempson20154}. The research within categorial grammar either suggests  that elliptic phenomena should be treated using a specific controlled form of copying  of information at the antecedent and moving it to the site of ellipsis, e..g in  \cite{jager1998multi,morrill1996generalising}, or by maintaining a non-directional  functional type  (meaning that it is not sensitive to where its argument occurs, before or after it), which is backward/forward looking, e.g. in \cite{KubotaLevine2017,jager2006anaphora}.  The first proposal can also be implemented using different modal Lambek Calculi, e..g that  developed in \cite{Moortgat97} and the second one using Displacement Calculus \cite{morrill2010calculus}; Abstract Categorial Grammars of \cite{Muskens2003,deGroote2001}, which allow for a separation of syntax and semantics within a categorial grammar and  allow for freedom of copying and movement at the semantic side,  can also be employed. We will not go too much into philosophical discussion in this paper   and base our work on an extension of the Lambek calculus with a limited form of contraction (shorthanded to \textbf{LCC}) via a non-directional functional type, introduced by J{\"a}ger \cite{jager2006anaphora}. In a previous paper \cite{wijholdsSadr2017} we showed how one can treat ellipsis using a controlled form of copying and movement via contraction and a modality.  What was novel in previous work was  that we discovered and showed how the use of Frobenius copying/dispatching of information does not work for resolving ellipsis, as it cannot distinguish between the sloppy and strict readings.  
Similar to previous  work \cite{wijholdsSadr2017}, we argue for a simple revision of the DisCoCat framework \cite{coecke2013lambek} in order to allow us to incorporate a proper notion of reuse of resources: instead of directly interpreting derivations as linear maps, where it becomes impossible to have a map that copies word embeddings \cite{jacobs2011bases,abramsky2009no}, we decompose the interpretation of grammar derivations into a two-step process, relying on a non-linear simply typed lambda calculus, in the style of \cite{muskens2016context,muskens2019context}. In doing so, we obtain a model that allows for the reuse of embeddings, while staying in the realm of vector spaces and linear maps. The novel part of the current paper, apart from its use of a backward looking bidirectional operation in Lambek Calculus,  rather than copying and moving syntactic information around, is that  we test our hypothesis on a well known verb disambiguation task \cite{mitchell2008vector,grefenstette2011experimental,kartsaklis2013prior}  using vectors and matrices obtained from large scale data. 

The paper is structured as follows: section \ref{sec:prelim} discusses the problem of ellipsis and anaphora and argues for non-linearity in the syntactic process, section \ref{sec:typelogical} gives the general architecture of our system. We proceed in section \ref{sec:deriving_ellipsis} with our main analysis and carry out a simple experiment in section \ref{sec:experiment} to show how our model may be empirically validated. We conclude with a discussion and future work in section \ref{sec:conclusion}.

\section{Ellipsis and Non-Linearity}
\label{sec:prelim}

Ellipsis can be defined as a linguistic phenomenon in which the full content of a sentence differs from its representation. In other words, in a case of ellipsis a phrase is \emph{missing} some part needed to recover its meaning. There are numerous types of ellipsis with a varying degree of complexity, but we will stick with verb phrase ellipsis, in which very often an ellipsis \emph{marker} is present to mark what part of the sentence is missing and where it has to be placed.

An example of verb phrase ellipsis is in Eq. \ref{ex:simple_ellipsis}, where the elided verb phrase is marked by the auxiliary verb. Ideally, sentence \ref{ex:simple_ellipsis}(a) is in a bidirectional entailment relation with \ref{ex:simple_ellipsis}(b), i.e. (a) entails (b) and (b) entails (a).

\begin{equation}\label{ex:simple_ellipsis}
\begin{tabular}{cll}
$a$ & \quotes{Alice drinks and Bill does too}\\
$b$ & \quotes{Alice drinks and Bill drinks}\\
\end{tabular}
\end{equation}
More complicated examples of ellipsis introduce an ambiguity; the example in Eq. \ref{ex:ellipsis_anaphora}, has a sloppy (b) and a strict (c) interpretation for (a).
\begin{equation}\label{ex:ellipsis_anaphora}
\begin{tabular}{cll}
$a$ & \quotes{Gary loves his code and Bill does too} & (ambiguous)\\
$b$ & \quotes{Gary loves Gary's code and Bill loves Bill's code} & (sloppy)\\
$c$ & \quotes{Gary loves Gary's code and Bill loves Gary's code} & (strict)\\
\end{tabular}
\end{equation}
In a formal semantics account, the first example could be analysed with the auxiliary verb as an identity function on the main verb of the sentence and an intersective meaning for the coordinator. Somehow the parts need to be appropriately combined to produce the reading (b):

\begin{center}
\begin{tabular}{lll}
	\begin{tabular}{c}
		doestoo : $\abs{x}{x}$ \\
		and : $\abs{x}{\abs{y}{(x \wedge y)}}$
	\end{tabular}
	& should give & $\mathbf{drinks}(\mathbf{alice}) \wedge \mathbf{drinks}(\mathbf{bill})$
\end{tabular}
\end{center}
The second example would assume the same meaning for the coordinator and auxiliary but now the possessive pronoun \quotes{his} gets a more complicated term: $\abs{x}{\abs{y}{\mathbf{owns}(x,y)}}$. The analysis then somehow should derive two readings: \\
	$$\mathbf{loves}(\mathbf{gary},x) \wedge \mathbf{owns}(\mathbf{gary},x) \wedge \mathbf{loves}(\mathbf{bill},x)$$
	$$\mathbf{loves}(\mathbf{gary},x) \wedge \mathbf{owns}(\mathbf{gary},x) \wedge \mathbf{loves}(\mathbf{bill},y) \wedge \mathbf{owns}(\mathbf{bill},y)$$

Indeed, these are meanings that would be produced by the approach of Dalrymple et al. \cite{dalrymple1991ellipsis}.

There are three issues with these analyses that are not solved in current distributional semantic frameworks: first, it is unclear what the composition operator is that maps from the meanings of the words to the meaning of the phrases. Second, it is unclear how the lexical constants (mainly the intersection operation expressed by the conjunction $\wedge$) are to be interpreted as a linear map. Thirdly, these examples contain a non-linearity; resources may be used more than once (the main verb is used twice in the first example, the noun phrases in the second example). We will outline a model that deals with all three.

The challenge of composition will be treated by using a compositional distributional semantic model in the lines of Coecke et al. \cite{coecke2013lambek}. For the interpretation on the lexical level, we will make use of Frobenius algebras to specify the lexical meaning of the coordinator and relative pronoun, following Kartsaklis \cite{kartsaklis2016coordination} and  Sadrzadeh et al. \cite{sadrzadeh2013frobenius}, respectively. Moreover, we will assign a similar meaning to the possessive pronoun.

What remains is to decide how to deal with non-linearity. With non-linearity we mean the possible duplication of resources, not the use of non-linear maps. On the side of vector semantics, though, it can be tempting to rely on the use of Frobenius algebras and use their dispatching operation to deal with copying, and indeed this operation has been  referred to as copying in the literature, e.g. see  \cite{CoeckePaq,CoeckeVic,kartsaklis2016coordination,sadrzadeh2013frobenius}.  Going a bit deeper, however, reveals that this operation places a vector into the diagonal of a matrix, that is, for a finite dimensional vector space $W$ spanned by basis $\{\overrightarrow{b_i}\}_i$, we have 
\[
\Delta \colon W \to W \otimes W \quad \mbox{given by} \quad \Delta (\sum_i C_i \overrightarrow{b_i}) = \sum_{ii} C_i \overrightarrow{b_i} \otimes \overrightarrow{b_i}
\]
As an example consider a two dimensional space $W$. A vector in this space will be copied via $\Delta$ into a matrix in $W \otimes W$, whose diagonals are $a$ and $b$ and whose non-diagonals are 0:
\[
\left(\begin{array}{c}a\\b\end{array}\right) \stackrel{\Delta}{\to}
\left(\begin{array}{cc}a & 0\\0& b\end{array}\right)
\]
 This is computed from the  definition of $\Delta$ on the basis $\Delta(\overrightarrow{b_i}) = \overrightarrow{b_i} \otimes \overrightarrow{b_i}$ and the fact that $\Delta$ is linear. Indeed it seems that $\Delta$ only ``copies'' the   basis into their tensors and it has  been shown that any other form of copying in this context, e.g. a Cartesian one,  is not allowed, see
\cite{jacobs2011bases,abramsky2009no} for proofs.  For a concrete linguistic demonstration of this fact, consider the anaphoric sentence  \quotes{Alice loves herself}, with the following noun vector and verb tensor:
	$$\mathbf{alice} = \sum\limits_{i} a_i \vec{v}_i \qquad \mathbf{loves} = \sum\limits_{jkl} c_{jkl} (\vec{v}_j \tensor \vec{s}_k \tensor \vec{v}_l)$$
We want to obtain the interpretation
$$\mathbf{alice}_i \mathbf{loves}_{ijk} \mathbf{alice}_k = \sum\limits_{ijk} a_i c_{ijk} a_k \vec{s}_{j}$$
However, using Frobenius copying would not give the desired result but rather something else:
$$\mathbf{alice}_i \mathbf{loves}_{iji} = \sum\limits_{ij} a_i a_i c_{iji} \vec{s}_j$$

Note that in the above, we use a simplification of  Einstein's  index notation for tensors. In Einstein notation, a tensor has indices on the top and bottom, specifying which index refers to a row or a column. For instance, a matrix is denoted by $^iM_j$, when $i$ enumerates the row elements and $j$ the column elements. We, however, only work  with finite dimensional vector spaces where a space is isomorphic to its dual space. In such cases, the Einstein notation simplifies and one can write  both of the subscripts under (or above). 

The use of Frobenius operations in itself may not immediately appear to be a problem, but in previous work \cite{wijholdsSadr2017} we show how a categorical model in the framework of Coecke et al. \cite{coecke2013lambek}, using Frobenius operations as a 'copying' operation, and treating the auxiliary phrase \quotes{does too} as an identity map, is not able to distinguish between the sloppy versus strict interpretations of a sentence with an elliptical phrase such as ``Alice loves her code, so does Mary". The  vector interpretations of both of the cases (1) ``Alice loves Alice's code and Mary loves Alice's code" and (2) ``Alice loves Alice's code and Mary loves Mary's code" become the following expression:

$$\mathbf{Alice} \tensor \mathbf{loves} \tensor \mathbf{her} \tensor \mathbf{code} \tensor \mathbf{and} \tensor \mathbf{Mary} \tensor \mathbf{does \ too} \mapsto \Delta_N(\mathbf{Alice} \odot \mathbf{Mary} \odot \mathbf{code})_{ik} \mathbf{loves}_{ijk}$$ 

Even though one may attempt to fix this problematic behaviour by complicating the meaning of an auxiliary verb, it shows that under reasonable assumptions a direct translation of proofs into a Frobenius tensor based model is not desirable for approaches that require this non-linear behaviour. In order to still have Cartesian copying behaviour in a tensor based model, we decompose the DisCoCat model of \cite{coecke2013lambek} into a two-step architecture: we first define an extension of the Lambek calculus, which allows for limited contraction, developed in \cite{jager2006anaphora}. In this setting, grammatical derivations as well as lexical entries are interpreted in a non-linear simply typed $\lambda$-calculus. The second stage of interpretation homomorphically maps these abstract meaning terms to terms in a lambda calculus of  vectors,  tensors,  and linear maps, developed in \cite{muskens2016context}. The final effect is that we allow the Cartesian behavior of copying elements before concretisation in a vector semantics: the meaning of a sentence now is a program that has non-linear access to word embeddings.

\section{Typelogical Distributional Semantics}
\label{sec:typelogical}

In a very general setting, compositionality can be defined as a homomorphic image (or functorial passage) from a syntactic algebra (or category) to a semantic algebra (or category). The only condition, then, is that the semantic algebra be weaker than the syntactic algebra: each syntactic operation needs to be interpretable by a semantic operation. To give a formal semantic account one would map the proof terms of a categorial grammar, or rewritings of a generative grammar, to the semantic operations of abstraction and application of some lambda term calculus. In a distributional model such as the one of Coecke et al. \cite{coecke2010mathematical,coecke2013lambek}, derivations of a Lambek grammar are interpreted by linear maps on finite dimensional vector spaces. For our presentation it will suffice to say that the Lambek calculus can be considered to be a \emph{monoidal biclosed category}, which makes the mapping to the compact closure of vector spaces straightforward. However, we want to employ the copying power of non-linear lambda calculus, and so we will move from the direct interpretation below \\

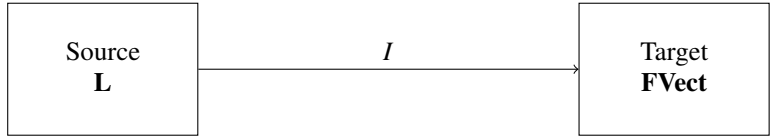
\begin{center}
\begin{tikzpicture}
	\node[draw, minimum width=2.5cm, minimum height=1.75cm] (der) 
	{\begin{tabular}{c}
	Source\\
	$\mathbf{L}$\\
	\end{tabular}};
	\node[draw, right=5cm of der, minimum width=2.5cm, minimum height=1.75cm] (sem)
	{\begin{tabular}{c}
	Target\\
	$\mathbf{FVect}$\\
	\end{tabular}};
	\draw[->] (der) edge node[above] {$I$} (sem);			
\end{tikzpicture}
\end{center}

to a two-step interpretation process: \\

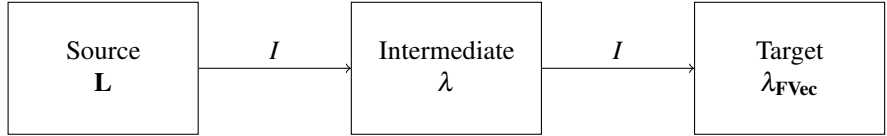
\begin{center}
\begin{tikzpicture}
	\node[draw, minimum width=2.5cm, minimum height=1.75cm] (der) 
	{\begin{tabular}{c}
	Source\\
	$\mathbf{L}$\\
	\end{tabular}};
	\node[draw, right=2cm of der, minimum width=2.5cm, minimum height=1.75cm] (sem1)
	{\begin{tabular}{c}
	Intermediate\\
	$\mathbf{\lambda}$\\
	\end{tabular}};
	\node[draw, right=2cm of sem1, minimum width=2.5cm, minimum height=1.75cm] (sem2)
	{\begin{tabular}{c}
	Target\\
	$\mathbf{\lambda_{FVec}}$\\
	\end{tabular}};
	\draw[->] (der) edge node[above] {$I$} (sem1);
	\draw[->] (sem1) edge node[above] {$I$} (sem2);			
\end{tikzpicture}
\end{center}

What we end up with is in fact a more intricate target than in the direct case: target expressions are now lambda terms with a tensorial interpretation, i.e. a program with access to word embeddings. The next subsections outline the details: we consider the syntax of the Lambek calculus with limited contraction, the semantics of non-linear lambda calculus, and the interpretation of lambda terms in a lambda calculus with tensors.

\subsection{Derivational Semantics: Formulas, Proofs and Terms}
\label{subsec:formulasproofs}

We start by introducing the Lambek Calculus with Limited Contraction \textbf{LLC}, a conservative extension of the Lambek calculus \textbf{L}, as defined in J{\"a}ger's monograph \cite{jager2006anaphora}.

\textbf{LLC} was in first instance defined to deal with anaphoric binding, but has many more applications including verb phrase ellipsis and ellipsis with anaphora. The system extends the Lambek calculus with a single binary connective $|$ that behaves like an implication for anaphora: a formula $A|B$ says that a formula $A$ can be bound to produce formula $B$, while retaining the formula $A$. This non-linear behaviour allows the kind of resource multiplication in syntax that one expects when dealing with anaphoric binding and ellipsis.

More formally, formulas or types of \textbf{LLC} are built given a set of basic types $T$ and using the following definition:

\begin{definition} Formulas of \textbf{LLC} are given by the following grammar, where $T$ is a finite set of basic formulas:
$$ A,B := T \ | \ A \bullet B \ | \ A \bs B \ | \ B \s A \ | \ A|B $$
\end{definition}

Intuitively, the Lambek connectives $\bullet, \bs, \s$ represent a \singlequotes{logic of concatenation}: $\bullet$ represents the concatenation of resources, where $\bs, \s$ represent directional decatenation, behaving as the residual implications with respect to the multiplication $\bullet$. The extra connective $|$ is a separate implication that behaves non-linearly: its deduction rules allow a mix of permutations and contractions, which effectively treat anaphora and VP ellipsis markers as phrases that look leftward to find a proper binding antecedent. Our convention is that we read $A | B$ as a function with input of type $A$ and an output of type $B$. 

\begin{figure}[h]
\begin{center}
\[\infer[Lex]{x : \sigma(w)}{w} \] \\
\begin{tabular}{c@{\hskip 3em}c}
\infer[I \bullet]{\langle M, N \rangle : A \bullet B}{M : A & N : B}
&
\infer[E \bullet]{\pi_1(M) : A \qquad \pi_2(M) : B}{M : A \bullet B} \\
& \\
& \\
\infer[I \bs, i]{\lambda x.M : A \bs B}{\infer[]{\quad M : B \quad}{\infer*[]{}{\infer[i]{x : A}{}} & \infer*[]{}{}}}
&
\infer[E \bs]{M \ N : B}{N : A & M : A \bs B} \\
& \\
& \\
\infer[I \s, i]{\lambda x.M : B \s A}{\infer[]{\quad M : B \quad}{\infer*[]{}{} & \infer*[]{}{\infer[i]{x : A}{}}}}
&
\infer[E \s]{M \ N : B}{M : B \s A & N : A} \\
& \\
\end{tabular}
\begin{center}
\begin{tabular}{m{3em}m{0.25em}m{2em}}
	$[N : A]_i$ & ... & $\infer[E |,  i]{M\ N : B}{M : A | B}$
\end{tabular}	
\end{center}
\end{center}
\caption{Labelled natural deduction for \textbf{LLC}.}
\label{nldia}
\end{figure}

The rules of  \textbf{LLC} are given in a natural deduction style  in Figure \ref{nldia}. The $Lex$ rule is an axiom of the logic: it allows us to relate  the judgements  of the logic to the words of the lexicon.  For instance, in the example proof tree provided in Figure \ref{fig:simpleder}, the judgement $\texttt{alice} : np$ is related to the word \emph{Alice}, the judgement $\texttt{bob} : np$ to the word \emph{Bob}, and the judgement $\texttt{and} : (s \bs s) / s$ to the word \emph{and}. Then, as it is usual in natural deduction, every connective has an introduction rule, marked with $I$ and an elimination rule, marked with $E$. In the introduction rules for $/$ and $\bs$,  the variable $x$ stands for an axiom,  in the introduction rule for $\bullet$ and eliminations rules for  $\bullet, /$ and $\bs$, we  have proofs for the premise types $A, B, A \bullet B, A/B$ and $A\bs B$, i.e. general terms $N$ and  $M$.

Informally speaking,  the introduction rule for $\bullet$, takes two terms $M$ and $N$, one of which ($M$) proves a formula $A$ and another of which ($N$) proves the formula $B$, and it pairs the terms with the tensor product of the formulae. that is, tells us that the $\langle M,N\rangle$ proves $A \bullet B$. The elimination rule for $\bullet$ takes a pair of terms, denoted by $M$ and tells us that the first projection of $M$, i.e. $\pi_1(M)$, i.e. the first element of the pair, proves $A$ and its second projection/element proves $B$. The introduction rule for $\bs$ takes the index of the rule where formula $A$ was proved using a term $x$, a proof tree which used this rule and possibly together with some other rules proved the formula $B$ using the term $M$, then derives  the formula $A\bs B$ using the lambda term $\lambda x. M$. The lambda terms are explained later on, but for now, think of this term as a function $M$  with the variable $x$. The elimination rule for $\bs$ is doing the opposite of what we just explained and which is what the introduction rule did. It takes a term $x$ for formula $A$, a term $y$ for formula $A \bs B$, then tells us that we can apply $y$ to $x$ to get something of type $B$. The rules for $/$ are similar to these but with different ordering, which is easily checkable from their proof rules in Figure  \ref{nldia}.

That brings us to the main rules that differentiate \textbf{LLC} from \textbf{L}  (the Lambek Calculus): the rules for $|$. Here, the elimination rule tells us that if somewhere in the proof we had proved  $A$ from $N$, and denoted the result by an index $i$,  and  then later  we encounter a term $M$ for $A | B$, and that $i$  happened before $M \colon A | B$, then we are allowed to eliminate  $|$  and get $B$ by applying the term $M$ to the term $N$. This rule is very similar to either of the $\bs$ and $/$ rules, in that it says you can eliminate the connective by applying its term to the term of one of its compartments, i.e. its input. The exception for the $|$ elimination rule is that it allows for that input, i.e. $[N : A]_i $ to happen not directly as the antecedent of the elimination rule, but as one of the other rules in the proof, somewhere before the current elimination rule. 
%
%
We can see how this rule is applicable in the proof tree of Figure  \ref{fig:simpleder}: we see a proof for $[\texttt{drinks} : np\bs s$, in this occasion indexed with a label $i$, then quite later on in the proof (actually at the end of it), we encounter the term $\texttt{dt drinks}: np\bs s$, now the $E|,i$ rule allows us to apply the latter to the former, all the way back, to obtain $\texttt{dt}: (np \bs n) | (np \bs s)$.   The $|$ connective also has an introduction rule, a proper formulation of this rule, however,  is more delicate. Since our anaphoric expressions are already  typed in the lexicon,  we do not need this rule in our paper and  refer the reader for different formulations and explanations of it to  J{\"a}ger's  book \cite[pp.123--124]{jager2006anaphora}.


The interpretation of proofs is established by a non-linear simply typed lambda term calculus, which labels the natural deduction rules of the calculus:

\begin{definition} Given a countably infinite set of variables $V = \{x,y,z...\}$ , terms of $\mathbf{\lambda}$ are as in the below grammar:
	$$M,N := V \ | \ \lambda x. M \ | \ M \ N \ | \ \langle M,\ N \rangle \ | \ \pi_1(M) \ | \ \pi_2(M)$$
\end{definition}

Terms obey the standard $\alpha$-, $\eta$- and $\beta$-conversion rules:

\begin{definition} For terms of $\mathbf{\lambda}$ we define three conversion relations:
	\begin{center}
	\begin{enumerate}
		\item $\alpha$-conversion: for any term $M$ we have
				\begin{center}
				\begin{tabular}{ccc}
				$M$ & $=_{\alpha}$ & $M[x \mapsto y]$
				\end{tabular}
				\end{center}
			provided that $y$ is a fresh variable, i.e. it does not occur in $M$. \\
			
		\item $\eta$-conversion: for terms $M$ we have \\
				\begin{center}
				\ra{1.2}
				\begin{tabular}{lcl}
					$\lambda x. M\ x$ & $=_{\eta}$ & $M \quad \text{($x$ does not occur in $M$)}$ \\
					$\langle \pi_1(M),\ \pi_2(M) \rangle$ & $=_{\eta}$ & $M$ \\
				\end{tabular}
				\end{center}			
				
		\item $\beta$-conversion: for terms $M$ we define
				\begin{center}
				\begin{tabular}{ccc}
					$(\lambda x.M)\ N$ & $\rightarrow_{\beta}$ & $M[x \mapsto N]$ \\
					$\pi_1(\langle M,\ N \rangle)$ & $\rightarrow_{\beta}$ & $M$ \\
					$\pi_2(\langle M,\ N \rangle)$ & $\rightarrow_{\beta}$ & $N$ \\
				\end{tabular}
				\end{center}
				We moreover write $M \twoheadrightarrow_{\beta} N$ whenever $M$ converts to $N$ in multiple steps.
	\end{enumerate}
	\end{center}
\end{definition}

The full labelled natural deduction is given in Figure \ref{nldia}. Proofs and terms give the basis of the derivational semantics; given a lexical map  relation  $\sigma  \subseteq  \Sigma \times  F$ for $\Sigma$ a dictionary of words, we  say that a sequence of words $w_1,...,w_n$ derives the formula $A$ whenever it is possible to derive a term $M:A$ with free variables $x_i$ of type $\sigma(w_i)$. For the $x_i$, one can substitute constants $c_i$ of type $\sigma(w_i)$ representing the meaning of the actual words $w_1,...,w_n$. The \emph{abstract meaning} of the sequence is thus given by the lambda term $t$. An example of such a proof is given for the elliptical phrase \quotes{Alice drinks and Bob does-too} in Figure \ref{fig:simpleder}. More involved examples are given in Figures \ref{fig:sloppyder} and \ref{fig:strictder}; they will be discussed  in section \ref{sec:deriving_ellipsis}.

\begin{figure}[h]
\infer[E \bs]{(\texttt{and} \ ((\texttt{dt} \ \texttt{drinks}) \ \texttt{bob})) (\texttt{drinks} \ \texttt{alice}) : s}{\infer[E \bs]{\texttt{drinks} \ \texttt{alice} : s}{\infer[Lex]{\texttt{alice} : np}{Alice} & \infer[Lex]{[\texttt{drinks} : np \bs s]_i}{drinks}} & \infer[E \s]{\texttt{and} \ ((\texttt{dt} \ \texttt{drinks}) \ \texttt{bob}) : s \bs s}{\infer[Lex]{\texttt{and} : (s \bs s) \s s}{and} & \infer[E \bs]{(\texttt{dt} \ \texttt{drinks}) \ \texttt{bob} : s}{\infer[Lex]{\texttt{bob} : np}{Bob} & \infer[E |, i]{\texttt{dt} \ \texttt{drinks} : np \bs s}{\infer[Lex]{\texttt{dt} : (np \bs s) | (np \bs s)}{does\ too}}}}}
\caption{Short hand derivation for \quotes{Alice drinks and Bob does-too}.} 
\label{fig:simpleder}
\end{figure}

\subsection{Lexical Semantics: Lambdas, Tensors and Substitution}
\label{subsec:interpretation}

We complete the vector semantics by adding the second step in the interpretation process, which is the insertion of lexical entries for the assumptions occurring in a proof. In this step,  we face the issue that interpretation directly into a vector space is not an option given that there is no copying map that is linear, while at the same time lambda terms don't seemingly reflect vectors. We solve the issue by showing, following \cite{muskens2016context,muskens2019context},  that vectors can be emulated using a lambda calculus.

\subsubsection{Lambdas and Tensors}

The idea of modelling tensors with lambda calculus is to represent vectors as functions from natural numbers to the values in the underlying field. This representation treats vectors as lists of elements of a field, for instance the field of reals $\reals$.  What the  function is doing is enumerating  the elements of this list. So for instance, consider  the following vector
\[
\overrightarrow{v} = [a,b,c, \dots] \qquad \text{for} \ a,b,c,{\dots} \in \reals
\]
The  representation of $\overrightarrow{v}$  using  a function $f$ becomes as follows
\[
f(1) = a,\ f(2) = b,\ f(3) = c, \dots \text{ and so on}
\]
For natural language applications, it is convenient to work with a fixed set of indices rather than  directly working with natural numbers as the starting point. These indices will be the ``context words" of the vector space model of word meaning.   For demonstration purposes, suppose these context words are the following set of words 
\[
C = \{\mbox{human, painting, army, weapon, marathon}\}
\]
Then a ``target word", i.e. the word whose meaning we are representing using these context words, will have values from $\reals$ in the entries of a vector spaces spanned by the above  context set.  For instance, consider three target words ``warrior", ``sword",  ad ``athlete". Their vector representations are as follows:
\begin{eqnarray*}
\overrightarrow{\text{warrior}} &=& [4,1,2,9,1]\\ 
\overrightarrow{\text{sword}} &=& [2,3,9,2,0]\\ 
\overrightarrow{\text{athlete}} &=& [6,2,0,1,9] 
\end{eqnarray*}
In a functional notation, our index set is the set of the context words, e.g. $C$, as given above, and for each target word $t$, our  function returns its value on each of the context words. So for instance, for a function $f$, the vector representation of ``warrior"  becomes as follows
\[
f(\text{human}) = 4, f(\text{painting}) = 1, f(\text{army}) = 2,  f(\text{weapon}) = 9, f(\text{marathon}) = 1
\]
Type-theoretically,  instead of working with a set of words as the domain of the representation function $f$, we enumerate the set of context words and use their indices as inputs to $f$. So we denote our set  $C$ above by indices $i_1, i_2, \dots, i_5$, which changes  the function representation to the following 
\[
f(i_1) = 4, f(i_2) = 1, f(i_3) = 2,  f(i_4) = 9, f(i_5) = 1
\]

That is, for any dimensionality $n$, we assume a basic type $I_n$, representing a finite index set (in concrete models the number of index types will be finite) of  context words. The underlying field, in the case  of natural language applications remains the set of real numbers $\reals$;   we denote  it by the type $R$. For more information about $\reals$ as a type, see \cite{muskens2016context}.  As explained above, the type of a vector in $\reals^n$ becomes  $V^n = I_n \rightarrow R$. Similarly,  the type of an $n \times m$ matrix, which is  vector in a space whose basis are pairs of words,  is $M^{n\times m} = I_n \rightarrow I_m \rightarrow R$. In general, we may represent an arbitrary tensor with dimensions $n,m,...,p$ by $T^{n\times m ... \times p} = I_n \rightarrow I_m \rightarrow ... \rightarrow I_p \rightarrow R$. We abbreviate cubes  $T^{n \times m \times p}$ to $C$  and hypercubes $T^{n \times m \times p \times q}$ to $H$.   We will leave out the superscripts denoting dimensionality when they are either irrelevant or understood from the context.

By reference to index notation for linear algebra, we write $v \ i$ as $v_i$ whenever it is understood that $i$ is of type $I$.
%
We moreover assume constants for the basic operations of a vector space: $0 : R, 1 : R, + : R \rightarrow R \rightarrow R, \cdot : R \rightarrow R \rightarrow R$ with their standard interpretation. Some standard operations can now be expressed using lambda terms: \\

\begin{center}
\ra{2}
\begin{tabular}{@{}ccc@{}}\toprule
	Name & Symbol & Lambda term \\ \midrule
	Matrix transposition & $T$ & $\lambda mij. m_{ji} : M \rightarrow M$ \\
	Matrix-Vector  multiplication & $\times_1$ & $\lambda mvi. \sum\limits_{j} m_{ij} \cdot v_j : M \rightarrow V \rightarrow V$ \\
	Cube-Vector  multiplication & $\times_2$ & $\lambda cvij. \sum\limits_{k} c_{ijk} \cdot v_k : C \rightarrow V \rightarrow M$ \\
	Hypercube-Matrix multiplication & $\times_3$ & $\lambda cmij. \sum\limits_{l} c_{ijkl} \cdot m_{kl} : H \rightarrow M \rightarrow M$ \\
	Vector Element wise multiplication & $\odot$ & $\lambda uvi. u_i \cdot v_i : V \rightarrow V \rightarrow V$ \\
	Vector addition & $+$ & $\lambda uvi. u_i + v_i : V \rightarrow V \rightarrow V$\\
	\bottomrule
\end{tabular}
\end{center}
\ \\
We can also express many other operations in the same, e.g. backwards matrix multiplication by composing matrix transposition with standard multiplication: $\times^T := \lambda mvi. \sum\limits_j m_{ji} \cdot v_j : M \rightarrow V \rightarrow V$. In the same way,  it is routine to  define a cube-matrix multiplication and a hypercube-cube and hypercube-vector multiplication. These operations do not occur in the current paper.  Similarly, one can define addition and element wise multiplication operations between matrices, cubes, and hypercubes. In what follows, we abuse the notation and denote the latter two with the same symbols, that is with $+$ and $\odot$ regardless of the type of object they are adding or multiplying. 

All of these operations, except for addition,  are instances of the multilinear algebraic operation of tensor contraction applicable to   any two tensors of arbitrary rank as long as they share at least one index. The  tensor contraction between them  is formed by applying the following formula:  
\begin{eqnarray*}
&&\sum_{i_1,...,i_{n+k}} A_{i_1 i_2 \cdots  i_n } B_{i_{n} i_{n+1} \cdots i_{n+k}}  \in \underbrace{W \otimes \cdots \otimes W}_{n+k-1}\\
&&\text{For}   \ \sum_{i_1,...,i_{n}} A_{i_1 i_2 \cdots i_n}\in \underbrace{W  \otimes \cdots \otimes W}_n  \quad \text{and} \quad  \sum_{i_n,...,i_{n+k}} B_{i_{n} i_{n+1} \cdots i_{n+k}} \in \underbrace{W \otimes \cdots \otimes W}_{k+1}
\end{eqnarray*}
Element wise multiplication between two vectors, or matrices, or tensors of the same rank is also an instance of tensor contraction, where one of the arguments of the multiplication is \emph{raised} to a tensor of a higher rank, with the argument in its diagonal and its other entries padded with zero. For an instance of this see \cite{kartsaklis2016coordination} where coordination is treated in a DisCoCat model, therein the author shows  how the linear algebraic closed form of  element wise multiplication arises as a result of a tensor contraction. 

\subsubsection{Lexical substitution}

To obtain a concrete model for a phrase, we need to  replace the abstract meaning term of a proof by a concrete tensor mapping. Since we map lambda terms to lambda terms, we only need to specify how constants $c$ are mapped to tensors. This will automatically induce a type-respecting term homomorphism $\cal{H}$. A general map that sends constants to a contraction friendly model is presented in Table \ref{table:tensor}.

\begin{table}[h]
\begin{center}
\ra{1.4}
\begin{tabular}{llll}\toprule
$w$&$\sigma(w)$ & ${\cal H}(w)$&${\cal T}(w)$\\ \midrule
\texttt{cn}& $n$ & \text{\bf cn}&$V$\\
\texttt{adj}& $n \s n$ & $\lambda v.(\text{\bf adj}\times_1 v)$&$VV$ \\
\texttt{adv}& $(np \bs s) \bs (np \bs s)$ & $\lambda m.(\text{\bf adv}\times_3 m)$&$HM$ \\
\texttt{itv}& $np \bs s$ & $\lambda v.(\text{\bf itv}\times_1 v)$& $VV$\\
\texttt{tv}& $(np \bs s) \s np$ & $\lambda uv.(\text{\bf tv}\times_2 u)\times_1 v$&$VVV$\\
\texttt{coord}& $(s \bs s) \s s$ & $\lambda P.\lambda Q. P \odot  Q$&$VVV$\\
\bottomrule
\end{tabular}
\end{center}
\caption{Translation that sends abstract terms to a tensor-based model using matrix and cube multiplication as the main operations; here an in the two other proceeding tables  the atomic types are $np$ and $s$. }
\label{table:tensor}
\end{table}

The different  composition operators of  Table \ref{table:tensor} seem to be different: we have matrix multiplication for adjectival phrases,  intransitive sentences and verb phrases,  cube multiplication  for transitive sentences, and pointwise multiplication for the conjunctive coordination.

Using  Table \ref{table:tensor} , we can translate the proof term of Figure \ref{fig:simpleder} as follows:
$$(\texttt{and} \ ((\texttt{dt} \ \texttt{drinks}) \ \texttt{bob})) (\texttt{drinks} \ \texttt{alice}) : s$$
and substitute the concrete terms to get the following $\beta$-reduced version:
$$\twoheadrightarrow_{\beta} (\mathbf{drinks} \times_1 \mathbf{alice}) \odot (\mathbf{drinks} \times_1 \mathbf{bob})$$

As another alternative, we can instantiate the proof terms in a multiplicative-additive model. This is a model where the  sentences are obtained by adding their individual word embeddings and the overall result is obtained by multiplying the two sentence vectors. This model is presented in Table \ref{table:additive}, according to which we  obtain the following semantics for our example sentence above:
$$\twoheadrightarrow_{\beta} (\mathbf{drinks} + \mathbf{alice}) \odot (\mathbf{drinks} + \mathbf{bob})$$
Another alternative is Table \ref{table:kronecker}, which provides the same terms with a Kronecker -based tensor semantics, originally used by \cite{grefenstette2011experimental} to model transitive sentences.

\begin{table}[h]
\begin{center}
\ra{1.4}
\begin{tabular}{llll}\toprule
$w$&$\sigma(w)$ & ${\cal H}(w)$&${\cal T}(w)$\\ \midrule
\texttt{cn}& $n$ & \text{\bf cn}&$V$\\
\texttt{adj}& $np \s n$ & $\lambda v.(\text{\bf adj} + v)$&$VV$ \\
\texttt{adv}& $(np \bs s) \bs (np \bs s)$ & $\lambda m.(\text{\bf adv} + m)$&$VV$ \\
\texttt{itv}& $np \bs s$ & $\lambda v.(\text{\bf itv} + v)$& $VV$\\
\texttt{tv}& $(np \bs s) \s np$ & $\lambda uv.(\text{\bf tv} + u + v)$&$VVV$\\
\texttt{coord}& $(s \bs s) \s s$ & $\lambda P.  Q. (P \odot Q)$&$VVV$\\
\bottomrule
\end{tabular}
\end{center}
\caption{Translation that sends abstract terms to a multiplicative-additive model.}
\label{table:additive}
\end{table}

We symbolise  the semantics of the basic elliptical phrase that comes out of any of these models for our example sentence as follows:
	$$M(\mathbf{sub}_1, \mathbf{verb}) \ \star \ M(\mathbf{sub}_2, N(\mathbf{verb}))$$
where $M$ is a general term for an intransitive sentence, $N$ is a term that modifies the verb tensor through the auxiliary verb, and $\star$ is an operation that expresses the coordination of the two subclauses. For a transitive sentence version, the above changes to the following:
	$$M(\mathbf{subj}_1, \mathbf{verb}, \mathbf{obj}_1) \ \star \ M(\mathbf{subj}_2, N(\mathbf{verb}), \mathbf{obj}_1)$$

Such a description is very general, and in fact allows us to derive almost all compositional vector models that have been tested in the literature (see e.g., \cite{milajevs2014evaluating}). This flexibility is necessary for ellipsis because it can model the Cartesian behaviour that is unavailable in a categorical modelling of vectors and linear maps.
Some models can, however,  only be incorporated by changing the lexical formulas associated to the individual words. The proposal of Kartsaklis et al \cite{kartsaklisverb} is one such example. They use the coordinator to a heavy extent and their typing and vector/tensor assignments  result in the following lambda semantics for the phrase \quotes{Alice drinks and Bob does too}: 
	$$\mathbf{drinks}  \times_1  (\mathbf{alice} \odot \mathbf{bob}) $$
The above is obtained by assigning an identity linear map to the auxiliary  phrase \singlequotes{does too} and then assigning a complex linear map to the coordinator \singlequotes{and} tailored in a way that it guarantees   the derivation of the final meaning. In our framework, we would need to take a similar approach, and we need to modify $M$  to essentially return the verb-subject pair, $N$ would be the identity, and \texttt{and} has to be defined with the tailored to purpose term below, which takes two pairs of subjects and verbs, but discards one copy of the verb to mimic the model of Kartsaklis et al \cite{kartsaklisverb}:
	$$\texttt{and} \qquad \lambda \langle s, v \rangle. \lambda \langle t, w \rangle. v \times_1 (s \odot t) $$

In either case, we can reasonably derive a large class of compositional functions that can be experimented with in a variety of tasks. With these tools in hand, we can give the desired interpretation to elliptical sentences in the next section.

\begin{table}[h]
\begin{center}
\ra{1.4}
\begin{tabular}{llll}\toprule
$w$&$\sigma(w)$ & ${\cal H}(w)$&${\cal T}(w)$\\ \midrule
\texttt{cn}& $n$ & \text{\bf cn}&$V$\\
\texttt{adj}& $np \s n$ & $\lambda v.(\text{\bf adj} \odot v)$&$VV$ \\
\texttt{adv}& $(np \bs s) \bs (np \bs s)$ & $\lambda m.(\text{\bf adv} \odot m)$&$VV$ \\
\texttt{itv}& $np \bs s$ & $\lambda v.(\text{\bf itv} \odot v)$& $VV$\\
\texttt{tv}& $(np \bs s) \s np$ & $\lambda uv.(\text{\bf tv} \odot (v \tensor u))$&$VVM$\\
\texttt{coord}& $(s \bs s) \s s$ & $\lambda P.\lambda Q. (P \odot Q)$&$VVV$\\
\bottomrule
\end{tabular}
\end{center}
\caption{Translation that sends abstract terms to a Kronecker model.  We abuse the notation to denote the element wise multiplication of two matrices with the same symbol, i.e. $\odot$, as the element wise multiplication of two vectors. }
\label{table:kronecker}
\end{table}

\section{Deriving Ellipsis: Strict and Sloppy Readings}
\label{sec:deriving_ellipsis}

In his book \cite{jager2006anaphora}, J{\"a}ger describes various applications of his logic \textbf{LLC}. With chapter 5 devoted to verb phrase ellipsis, he discusses various examples of general ellipsis: right node raising, gapping, stripping, VP ellipsis, antecedent contained deletion, and sluicing. Using these categories, an account is developed for VP ellipsis and sluicing. This treatment    directly carries over to the vectorial setting, with the challenge that  we need to think about how to fill in the lexical semantics.
We already gave the basic example of an elliptical phrase in Figure \ref{fig:simpleder}. In this section we show how the account of J{\"a}ger allows us to give compositional meanings to ellipsis with anaphora, and cascaded ellipsis, contrasting it with the direct categorical approach, which we show in \cite{wijholdsSadr2017} to be unsuitable for these cases.

\subsection{Ellipsis with Anaphora}

The interaction of ellipsis with anaphora leads to strict and sloppy readings, as already demonstrated in Section \ref{sec:prelim}. We repeat the example here and give the separate derivations:

\begin{equation}\label{ex:ellipsis_anaphora}
\begin{tabular}{cll}
$a$ & \quotes{Gary loves his code and Bob does too} & (ambiguous)\\
$b$ & \quotes{Gary loves Gary's code and Bob loves Bob's code} & (sloppy)\\
$c$ & \quotes{Gary loves Gary's code and Bob loves Gary's code} & (strict)\\
\end{tabular}
\end{equation}

The lexical assignment of type $np | (np \s n)$ to the possessive pronoun $his$ renders it an unbound anaphora, looking for a preceding noun phrase to bind to it. Similarly, the type assignment $(np \bs s) | (np \bs s)$ registers \singlequotes{does too} as the ellipsis marker that needs to be bound by a preceding verb phrase. The derivations of the strict (Figure \ref{fig:strictder}) and the sloppy (Figure \ref{fig:sloppyder}) readings essentially differ in their \emph{order of binding}: by binding \singlequotes{Gary} to the possessive pronoun and then binding the resulting verb phrase for \singlequotes{loves his code} to the ellipsis marker, we obtain the strict reading, whereas binding the verb phrase with the unbound possessive pronoun and subsequently binding the two copies of the pronoun differently, we get the sloppy reading. The flexibility of J{\"a}ger's approach is illustrated by the fact that one can  ultimately abstract  over the binding noun phrase to obtain a third reading, which would derive the type $np | s$, since that pronoun was left unbound.

\begin{figure}[h]
\infer[E \bs]{\deduce[]{s}{(v \ ((w \ (\lambda t.(y \ ((z \ t) \ u)) \ t)) \ w)) \ ((y \ ((z \ x) \ u)) \ x)}}{\hspace{0em} \infer[E \bs]{\deduce{s}{(y \ ((z \ x) \ u)) \ x}}{\infer[Lex]{\deduce{np}{x}}{Gary} & \infer[I \bs, 1]{\deduce{np \bs s}{[\lambda t.(y \ ((z \ t) \ u)) \ t]_j}}{\infer[E \bs]{\deduce{s}{(y \ ((z \ t) \ u)) \ t}}{\infer[1]{\deduce{np}{[t]_i}}{} & \infer[E \s]{\deduce{np \bs s}{y \ ((z \ t) \ u)}}{\infer[Lex]{\deduce{(np \bs s) \s np}{y}}{loves} & \infer[E \s]{\deduce{np}{(z \ t) \ u}}{\infer[E |, i]{\deduce{np \s n}{z \ t}}{\infer[Lex]{\deduce{np | (np \s n)}{z}}{his}} & \infer[Lex]{\deduce{n}{u}}{code}}}}}} \hspace{-7.5em} & \infer[E \s]{\deduce{s \bs s}{v \ ((w \ (\lambda t.(y \ ((z \ t) \ u)) \ t)) \ w)}}{\infer[Lex]{\deduce{(s \bs s) \s s}{v}}{and} & \infer[E \bs]{\deduce{s}{(w \ (\lambda t.(y \ ((z \ t) \ u)) \ t)) \ w}}{\infer[Lex]{\deduce{np}{w}}{Bob} & \infer[E |,j]{\deduce{np \bs s}{w \ (\lambda t.(y \ ((z \ t) \ u)) \ t)}}{\infer[Lex]{\deduce{(np \bs s)|(np \bs s)}{w}}{does\ too}}}}}
\caption{Sloppy interpretation for \quotes{Gary loves his code and Bob does-too}: Gary loves Gary's code and Bob loves Bob's code.}
\label{fig:sloppyder}
\end{figure}



\begin{figure}[h]
\infer[E \bs]{\deduce{s}{(v \ ((w \ (y \ ((z \ x) \ u))) \ w)) \ ((y \ ((z \ x) \ u)) \ x)}}{\infer[E \bs]{\deduce{s}{(y \ ((z \ x) \ u)) \ x}}{\infer[Lex]{\deduce{np}{[x]_i}}{Gary} & \infer[E \s]{\deduce{np \bs s}{[y \ ((z \ x) \ u)]_j}}{\infer[Lex]{\deduce{(np \bs s) \s np}{y}}{loves} & \infer[E \s]{\deduce{np}{(z \ x) \ u}}{\infer[E |, i]{\deduce{np \s n}{z \ x}}{\infer[Lex]{\deduce{np | (np \s n)}{z}}{his}} & \infer[Lex]{\deduce{n}{u}}{code}}}} \hspace{-2em} & \infer[E \s]{\deduce{s \bs s}{v \ ((w \ (y \ ((z \ x) \ u))) \ w)}}{\infer[Lex]{\deduce{(s \bs s) \s s}{v}}{and} & \infer[E \bs]{\deduce{s}{(w \ (y \ ((z \ x) \ u))) \ w}}{\infer[Lex]{\deduce{np}{w}}{Bob} & \infer[E |,j]{\deduce{np \bs s}{w \ (y \ ((z \ x) \ u))}}{\infer[Lex]{\deduce{(np \bs s)|(np \bs s)}{w}}{does\ too}}}}}
\caption{Strict interpretation for \quotes{Gary loves his code and Bob does-too}: Gary loves Gary's code and Bob loves Gary's code.}
\label{fig:strictder}
\end{figure}

If we assume a tensor-based compositional model that uses tensor contraction to obtain the meaning of a sentence, we get the two different meanings for the strict and sloppy readings as follows:

\begin{enumerate}
	\item $((\mathbf{loves} \times_2 (\mathbf{gary} \odot \mathbf{code})) \times_1  \mathbf{gary}) \odot ((\mathbf{loves} \times_2 (\mathbf{gary} \odot \mathbf{code}))  \times_1 \mathbf{bob}) \qquad \text{(strict)}$ \\
	\item $((\mathbf{loves} \times_2 (\mathbf{gary} \odot \mathbf{code}))  \times_1 \mathbf{gary}) \odot ((\mathbf{loves} \times_2 (\mathbf{bob} \odot \mathbf{code})) \times_1 \mathbf{bob}) \qquad \text{(sloppy)}$
\end{enumerate}

\subsection{Cascaded Ellipsis}

J{\"a}ger also describes the phenomenon of \emph{cascaded ellipsis}, in which an ellipsis contains an elided verb phrase within itself, as in \quotes{Gary revised his code before the student did, and Bob did too}. In this case there are three derivations possible (although even more readings could be found):

	\begin{enumerate}
		\item Gary revised Gary's code before the student revised Gary's code, and Bob revised Gary's code before the student revised Gary's code. \\[1em]	
				$\texttt{and}$ \\
				$(\texttt{before} \ (\texttt{revise} \ ((\texttt{his} \ \texttt{gary}) \ \texttt{code}) \ \texttt{student}) \ (\texttt{revise} \ ((\texttt{his} \ \texttt{gary}) \ \texttt{code}) \ \texttt{gary}))$ \\
				$(\texttt{before} \ (\texttt{revise} \ ((\texttt{his} \ \texttt{gary}) \ \texttt{code}) \ \texttt{student}) \ (\texttt{revise} \ ((\texttt{his} \ \texttt{gary}) \ \texttt{code}) \ \texttt{bob}))$ \\
		\item Gary revised Gary's code before the student revised Gary's code, and Bob revised Bob's code before the student revised Bob's code. \\[1em]
				$\texttt{and}$ \\
				$(\texttt{before} \ (\texttt{revise} \ ((\texttt{his} \ \texttt{gary}) \ \texttt{code}) \ \texttt{student}) \ (\texttt{revise} \ ((\texttt{his} \ \texttt{gary}) \ \texttt{code}) \ \texttt{gary}))$ \\
				$(\texttt{before} \ (\texttt{revise} \ ((\texttt{his} \ \texttt{bob}) \ \texttt{code}) \ \texttt{student}) \ (\texttt{revise} \ ((\texttt{his} \ \texttt{bob}) \ \texttt{code}) \ \texttt{bob}))$ \\
		\item Gary revised Gary's code before the student revised the student's code, and Bob revised Bob's code before the student revised the student's code. \\[1em]
				$\texttt{and}$ \\
				$(\texttt{before} \ (\texttt{revise} \ ((\texttt{his} \ \texttt{student}) \ \texttt{code}) \ \texttt{student}) \ (\texttt{revise} \ ((\texttt{his} \ \texttt{gary}) \ \texttt{code}) \ \texttt{gary}))$ \\
				$(\texttt{before} \ (\texttt{revise} \ ((\texttt{his} \ \texttt{student}) \ \texttt{code}) \ \texttt{student}) \ (\texttt{revise} \ ((\texttt{his} \ \texttt{bob}) \ \texttt{code}) \ \texttt{bob}))$ \\
	\end{enumerate}
	
A tensor-based model would assign three meanings appropriately. For example, the first subclause of 1 would give the following
	$$(\mathbf{gary} \times_1 \mathbf{revise} \times_2 (\mathbf{gary} \odot \mathbf{code}))\ \star \ (\mathbf{student} \times_1 \mathbf{revise} \times_2 (\mathbf{gary} \odot \mathbf{code}))$$
where $\star$ interprets the function word \singlequotes{before}.	

In the next section we carry out experimental evaluation of the framework developed so far. We start out with a toy experiment and then perform a large-scale experiment on verb phrase-elliptical sentences. We do not cover the more complex cases of ellipsis that involve ambiguities: setting up experiments for those cases is a task on its own and requires more investigation.

\section{Experimental Evaluation}
\label{sec:experiment}

To evaluate the framework we have developed so far, we carry out an experiment involving verb disambiguation. This kind of task was initiated in the work of Mitchell \& Lapata \cite{mitchell2008vector,mitchell2010composition} in order to evaluate the compositional vectors of intransitive sentences and verb phrases. These have been extended to transitive sentences Grefenstette \& Sadrzadeh and Kartsaklis \& Sadrzadeh \cite{grefenstette2011experimental,kartsaklis2013prior}. Here, we introduce the general idea behind the verb disambiguation task and how it is solved with compositional distributional models, before proceeding to an illustratory toy experiment and a large scale experiment.

A distributional model on the word level is considered successful if it optimises the similarity between words. Whenever two words $w_1$ and $w_2$ are considered similar, the associated vectors $\overrightarrow{w_1}$ and $\overrightarrow{w_2}$ ought to be similar as well. Similarity judgments between words are obtained by asking human judges, whereas the customary way of measuring similarity between vectors is given by the cosine of the angle between vectors (cosine similarity):
	$$\cos(\overrightarrow{v}, \overrightarrow{w}) = \frac{\overrightarrow{v} \cdot \overrightarrow{w}}{|\overrightarrow{v}|\ |\overrightarrow{w}|}$$
where $\cdot$ denotes the dot product and $| \cdot |$ denotes the magnitude of a vector.

Compositional tasks follow the same pattern, but now one is also interested in (a) how context affects similarity judgments and (b) how word representations are to be composed to give a sentence vector. The idea behind the verb disambiguation tasks \cite{mitchell2008vector,grefenstette2011experimental,kartsaklis2013prior} is that sentences containing an ambiguous verb can be disambiguated by context. An example  is the verb \emph{meet} which can mean \emph{visit} or  \emph{satisfy} (a requirement). In the sentence  \emph{Students  meet  teachers},  \emph{meet} means  \emph{visit}, whereas in the sentence  \emph{Houses meet standard}, it means  \emph{satisfy}. What makes this idea suitable for compositional distributional semantics is that we can use the vectors of these sentences to disambiguate the verb. This is detailed below.

Suppose we have a verb $V$ that is ambiguous between two different meanings $V_1$ and $V_2$, we refer to $V_1$ and $V_2$ as the two landmark meanings of $V$. We position $V$ in a sentence $Sbj \ V \ Obj$ in which only one of the meanings of the verbs makes sense. Suppose that meaning is $V_1$, so the sentences $Sbj \ V \ Obj$ and  $Sbj \ V_1 \ Obj$ make sense while the sentence $Sbj \ V_2 \ Obj$ does not. Then the cosine similarity between the vectors for the first two sentences ought to be high, but between those for the first and the third sentence it ought to be low. So the hypothesis that is tested is that this disambiguation by context manifests when we compute vectors for the meanings of these sentences. In technical terms, we wish the distance between the meaning vector of $Sbj \ V_1 \ Obj$  and that of $Sbj \ V \ Obj$ to be smaller than the distance between the vector of $Sbj \ V_2 \ Obj$  and that of $Sbj \ V \ Obj$:
\begin{eqnarray*}
\cos(\overrightarrow{Sbj \ V \ Obj},\ \overrightarrow{Sbj \ V_1 \ Obj}) &\geq& \cos(\overrightarrow{Sbj \ V \ Obj},\ \overrightarrow{Sbj \ V_2 \ Obj}) 
\end{eqnarray*}

This hypothesis forms the basis for our verb disambiguation task. Each of the datasets created for verb disambiguation \cite{mitchell2008vector,grefenstette2011experimental,kartsaklis2013prior} contains a balanced number of subjects, or subject-object combinations for several verbs and two landmark interpretations. That is, for a verb $V$ ambiguous between $V_1$ and $V_2$ there will be roughly an equal number of contexts that push the meaning of $V$ to $V_1$ and to $V_2$. Moreover, these datasets contain similarity judgments that allow us to not just classify the most likely interpretation of a given verb, but to compute the correlation between a model's prediction and the human judgments, to see how well a model aligns with humans.

The basic such models for composing word vectors to sentence vectors are the additive and multiplicative models that, for any sentence, simply add or multiply the vectors for the words in the sentence. For intransitive sentences of the form $Sbj\ V$, we would get respectively
		$$\overrightarrow{Sbj\ V} = \overrightarrow{Sbj} + \overrightarrow{V} \qquad \qquad \overrightarrow{Sbj\ V} = \overrightarrow{Sbj} \odot \overrightarrow{V}$$
For the transitive case, of the form $Sbj\ V\ Obj$, we additionally consider the Kronecker model used of \cite{grefenstette2011experimenting}, which assigns to a sentence \emph{subj verb obj} the following  formula:
	$$(\vek{verb} \tensor \vek{verb}) \odot (\vek{subj} \tensor \vek{obj})$$
In this model, note that the resulting representation is now a matrix rather than a vector.

Here, we extend the experimental setting to elliptical sentences, our hypothesis is twofold: on the one hand, an elliptical phrase will have more content that adds to the context of the verb to be disambiguated, allowing us to disambiguate more effectively. On the other hand, we test the disambiguating effect of resolving the ellipsis.
Going to an elliptical setting allows us to define several more composition models based on the additive/multiplicative and Kronecker models: for a transitive sentence $Sbj\ V\ Obj$ extended to the elliptical setting $Sbj \ V \ Obj \ and \ Sbj' \ does \ too$, we can again consider the additive and multiplicative models:
	$$\overrightarrow{Sbj \ V \ Obj \ and \ Sbj' \ does \ too} = \overrightarrow{Sbj} + \overrightarrow{V} + \overrightarrow{Obj} + \overrightarrow{and} + \overrightarrow{Sbj'} + \overrightarrow{does} + \overrightarrow{too}$$
	$$\overrightarrow{Sbj \ V \ Obj \ and \ Sbj' \ does \ too} = \overrightarrow{Sbj} \odot \overrightarrow{V} \odot \overrightarrow{Obj} \odot \overrightarrow{and} \odot \overrightarrow{Sbj'} \odot \overrightarrow{does} \odot \overrightarrow{too}$$
In addition, we can now consider combinations of models on the resolved elliptical phrases, following the pattern of Section 3. For an intransitive as well as a transitive sentence extended to an elliptical setting, its resolved version combines the two implicit subclauses by an operation. Hence, we can use one of the models outlined above on the subclauses, and then choose an operation to combine them. This leads, for the intransitive case, to the following four models: \\
\begin{center}
\ra{1.6}
\begin{tabular}{@{}ll@{}}\toprule
Model & Formula \\
\midrule
Multiplicative $\odot$ & $\vek{subj} \odot \vek{verb} \odot \vek{subj^*} \odot \vek{verb}$ \\
Multiplicative $+$ & $(\vek{subj} \odot \vek{verb}) + (\vek{subj^*} \odot \vek{verb})$ \\
Additive $\odot$ & $(\vek{subj} + \vek{verb}) \odot (\vek{subj^*} + \vek{verb})$ \\
Additive $+$ & $\vek{subj} + \vek{verb} + \vek{subj^*} + \vek{verb}$ \\
\bottomrule
\end{tabular}
\end{center}
For the transitive case, we additionally get the Kronecker $+$ and Kronecker $\odot$ models, given by either summing or multiplying the two Kronecker model matrices of the subclauses:
\begin{center}
\ra{1.6}
\begin{tabular}{@{}ll@{}}\toprule
Model & Formula \\
\midrule
Kronecker $+$ & $(\vek{verb} \tensor \vek{verb}) \odot (\vek{subj} \tensor \vek{obj}) + (\vek{verb} \tensor \vek{verb}) \odot (\vek{subj'} \tensor \vek{obj})$ \\
Kronecker $\odot$ & $(\vek{verb} \tensor \vek{verb}) \odot (\vek{subj} \tensor \vek{obj}) \odot (\vek{verb} \tensor \vek{verb}) \odot (\vek{subj'} \tensor \vek{obj})$ \\
\bottomrule
\end{tabular}
\end{center}

\subsection{A Toy Experiment}
In order to demonstrate the effect of vectors and distances in this task, we provide a hypothetical though intuitive example. Consider the sentence \quotes{the man runs}, which is ambiguous between \quotes{the man races} and \quotes{the man stands (for election)}. The sentence itself does not have enough context to help disambiguate the verb, but if we add a case of ellipsis such as  \quotes{the man runs, the dog too} to it, the ambiguity will be resolved. Another example, this time transitive,   is the sentence \quotes{the man draws the sword}, which is ambiguous between \quotes{the man pulls the sword} and \quotes{the man depicts the sword}. Again, the current sentence in which the ambiguous verb occurs may not easily disambiguate it, but after adding the extra context \quotes{the soldier does too}, the disambiguating effect of the context is much stronger.

Consider the following vector space built from   raw co-occurrence counts of several nouns and verbs with respect to a set of context words. The co-occurrence matrix is given in Table \ref{table:coocc}; each row of the table represents a word embedding.
\begin{table}[h]
\centering
\begin{tabular}{@{}ccccccc@{}}\toprule
 	& human & painting & army & weapon & marathon & election \\
\midrule
man & 2 & 3 & 4 & 2 & 4 & 4 \\
painter & 3 & 8 & 1 & 3 & 1 & 1 \\
warrior & 4 & 1 & 2 & 9 & 1 & 0 \\
sword & 2 & 3 & 9 & 2 & 0 & 0 \\
picture & 1 & 20 & 0 & 1 & 1 & 1 \\
governor & 7 & 1 & 1 & 3 & 1 & 9 \\
athlete & 6 & 2 & 0 & 1 & 9 & 1 \\
\midrule
draw & 4 & 10 & 9 & 11 & 2 & 3 \\
pull & 7 & 2 & 10 & 15 & 1 & 1 \\
depict & 3 & 15 & 2 & 2 & 1 & 2 \\
\midrule
run & 4 & 0 & 2 & 1 & 8 & 7 \\
race & 8 & 0 & 0 & 3 & 10 & 3 \\
stand & 5 & 1 & 0 & 1 & 2 & 11 \\

\bottomrule
\end{tabular}
\captionsetup{justification=centering}
\caption{(Hypothetical) co-occurrence counts for several nouns and verbs.}
\label{table:coocc}
\end{table}

%
We work out the cosine similarity scores between vector representations of a sentence with an ambiguous verb and its two landmark intepretations, following the models outlined above, on the concrete sentence \quotes{the man runs} with the extension of \singlequotes{governer} and \singlequotes{athlete} respectively. The idea is that the representation of \quotes{the man runs and governor does too} will be closer to that of \quotes{the man stands and governor does too}, whereas the representation of \quotes{the man runs and athlete does too} will be closer to that of \quotes{the man races and athlete does too}. The cosine similarity scores for each model are presented in Table \ref{table:intranssim}. The original representation of \quotes{the man runs} is more similar to \singlequotes{the man races} by a difference of $0.10$. However, for all models except the fully additive model, we see that adding the extra subject increases the difference between similarity scores, thereby making it easier to distinguish the correct interpretation. The most discriminative model is the fully multiplicative one, which treats the conjunctive coordinator as multiplication.

\begin{table}[h]
\centering
\begin{tabular}{@{}lcccccccc@{}}
\toprule
 & \multicolumn{2}{c}{Multiplicative} & \multicolumn{2}{c}{Multiplicative}   & \multicolumn{2}{c}{Additive} & \multicolumn{2}{c}{Additive}  \\
 & \multicolumn{2}{c}{$\odot$} & \multicolumn{2}{c}{$+$} & \multicolumn{2}{c}{$\odot$} & \multicolumn{2}{c}{$+$} \\ 
 	\cmidrule(lr){2-3} \cmidrule(l){4-5} 	\cmidrule(lr){6-7} \cmidrule(l){8-9}
 & \textbf{race} & \textbf{stand}  & \textbf{race} & \textbf{stand}  & \textbf{race} & \textbf{stand}  & \textbf{race}  & \textbf{stand}  \\
 \midrule
man run & .88 & .78 & .88 & .78 & .94 & .92 & .94 & .92 \\
man run, governor does too & \textbf{.47} & \textbf{.99} & .80 & .94 & .82 & .89 & .95 & .93 \\
man run, athlete does too & \textbf{.99} & \textbf{.36} & .96 & .71 & .94 & .71 & .95 & .92 \\
\bottomrule
\end{tabular}
\captionsetup{justification=centering}
\caption{Cosine similarity scores between representations involving the intransitive verb \singlequotes{run}. Column \textbf{race}: the representation of the corresponding row sentence but with \singlequotes{race} instead of \singlequotes{run}, similarly for \textbf{stand}.}
\label{table:intranssim}
\end{table}

For the transitive case, we compare the sentences \quotes{the man draws the sword} and \quotes{the man draws the picture} with their landmark interpretations in which the verb \singlequotes{draw} is replaced by either \singlequotes{pull} or \singlequotes{depict}. All of these are extended with the contexts \singlequotes{warrior} and \singlequotes{painter}, and we compute the result of four of the mixed transitive models outlined above for the elliptical case: two are the same additive models that just sum all the vectors in a (sub)clause and either sum or multiply vectors for the subclauses for the elliptical variant, and two models use the Kronecker representation detailed above. The concrete cosine similarity scores are displayed in Table \ref{table:transsim}. 

%
%
\begin{table}[h]
\centering
\begin{tabular}{@{}lcccccccc@{}}
\toprule
 & \multicolumn{2}{c}{Kronecker} & \multicolumn{2}{c}{Kronecker}   & \multicolumn{2}{c}{Additive} & \multicolumn{2}{c}{Additive}  \\
 & \multicolumn{2}{c}{$\odot$} & \multicolumn{2}{c}{$+$} & \multicolumn{2}{c}{$\odot$} & \multicolumn{2}{c}{$+$} \\ 
 	\cmidrule(lr){2-3} \cmidrule(l){4-5} 	\cmidrule(lr){6-7} \cmidrule(l){8-9}
 & \textbf{pull} & \textbf{depict}  & \textbf{pull}  & \textbf{depict}  & \textbf{pull}  & \textbf{depict}  & \textbf{pull}  & \textbf{depict} \\
 \midrule
man draw sword & .83 & .50 & .83 & .50 & .96 & .93 & .96 & .93 \\
man draw sword, warrior does too & \textbf{.98} & \textbf{.07} & .92 & .44 & .94 & .76 & .96 & .93 \\
man draw sword, painter does too & \textbf{.37} & \textbf{.28} & .69 & .59 & .89 & .80 & .96 & .93 \\
\midrule
man draw picture & .82 & .74 & .82 & .74 & .97 & .95 & .97 & .95 \\
man draw picture, warrior does too & \textbf{.98} & \textbf{.25} & .91 & .65 & .92 & .97 & .97 & .95 \\
man draw picture, painter does too & \textbf{.37} & \textbf{.95} & .68 & .88 & .96 & .98 & .97 & .96\\
\bottomrule
\end{tabular}
\captionsetup{justification=centering}
\caption{Cosine similarity scores between sentence representations using several models. Column \textbf{pull}: the representation of the corresponding row sentence but with \singlequotes{pull} instead of \singlequotes{draw}, similarly for \textbf{depict}.}
\label{table:transsim}
\end{table}
In this case, neither of the additive models  seem to be  effective: for the original phrases they already give very high similarity scores, and those do not change greatly after adding the extra context. For the Kronecker models, we see that the best discriminatory model is the one that multiplies the vectors for the subclauses: in both original transitive phrases the interpretation \singlequotes{pull} is more similar than \singlequotes{depict}, but adding the context improves the disambiguation results. For the first phrase, where a sword is drawn, the addition of \singlequotes{warrior} greatly improves the similarity with \singlequotes{pull} and accordingly decreases the similarity with \singlequotes{depict}, though for the addition of \singlequotes{painter} this is not the case. The representation for \singlequotes{painter} is in itself already closer to that of \singlequotes{depict} (cosine similarity of $0.97$) than it is to that of \singlequotes{pull} (cosine similarity of $0.52$), so adding \singlequotes{painter} to the sentence makes it harder to be certain about \singlequotes{pull} as a likely interpretation of \singlequotes{draw}. We see in fact that the difference between the two sentence interpretations has become smaller. 

For the second phrase, in which a picture is drawn, the original ambiguity is bigger, but adding the context provides us with the appropriate disambiguating scores. As with the first phrase, we also experience the difficulty in disambiguation: a human may deem \quotes{man draw picture, warrior does too} to be more similar to \quotes{man depict picture, warrior does too} since pulling a picture is not a very sensible action. However, because the vector for \singlequotes{warrior} is closer to that for \singlequotes{pull} (cosine similarity of $0.94$) than it is to \singlequotes{depict}  (cosine similarity of $0.31$) the model will favour the interpretation in which the picture is pulled.

\subsection{Large Scale Evaluation}
In addition to a hypothetical toy example, we experimented with our models on a large scale dataset, obtained by extending the disambiguation dataset of Mitchell and Lapata \cite{mitchell2008vector}, which we will refer to as the ML2008 dataset. The ML2008 is an instance of the verb disambiguation task that we have been discussing so far, and contains 120 pairs of sentences: For each of 15 verb triples ($V$, $V_1$, $V_2$), where verb $V$ is ambiguous between interpretation $V_2$ and $V_3$, four different context subjects were added, and the so constructed sentence pairs $Sbj V, Sbj V_1$ and $Sbj V, Sbj V_2$ were annotated for similarity by humans. For each of two sentence pairs, the interpretation that was assumed more likely was labelled HIGH before collecting annotations, and the other one was labelled LOW; this was done both for verification purposes as well as randomisation of the presentation of the sentence pairs to human judges. The subjects that were added would be mixed: some would cause the verb to tend to one interpretation, others cause the verb to be interpreted with the second meaning.

For example, the dataset contains the pairs

\begin{center}
	\begin{tabular}{lll}
		Landmark & HIGH & LOW \\
		\midrule
		export boom & export prosper & export thunder \\
		gun boom & gun thunder & gun prosper \\
		\midrule
	\end{tabular}
\end{center}

To extend such pairs to an elliptical setting, we chose a second subject for each sentence, as follows: for a given subject/verb combination and its two interpretations, we chose a new subject that occurred frequently in a corpus\footnote{In our case, this was the combined UKWaC and Wackypedia corpus, availabe at \href{http://wacky.sslmit.unibo.it}{\nolinkurl{wacky.sslmit.unibo.it}}}, but significantly more frequently with the more likely unambiguous verb (the one marked HIGH). For example, the word \quotes{economy} occurs with \quotes{boom} but it occurs significantly more often with \quotes{prosper} then it does with \quotes{thunder}. And similarly, \quotes{cannon} occurs with \quotes{boom} and \quotes{thunder} but not so often with \quotes{prosper}. We then format the pairs from the ML2008 dataset using the new subject and the elliptical setting. For the examples above, we then got
\begin{center}
	\begin{tabular}{cll}
		Landmark & export boom and economy does too  & gun boom and cannon does too \\
		\midrule
		HIGH &	   export prosper and economy does too & gun thunder and cannon does too \\
		LOW & 	   export thunder and economy does too & gun prosper and cannon does too \\
		\midrule
	\end{tabular}
\end{center}
In total, we added two new subjects to each sentence pair, producing a dataset of 240 entries. We used the human similarity judgments of the original ML2008 dataset to see whether the addition of disambiguating context, combined with our ellipsis model, will be able to better distinguish verb meaning. As explained in the start of this section, we use several different concrete models to compute the representation of the sentences in the dataset, and compute the cosine similarity between sentences in a pair; the predicted judgments are then evaluated by computing the (linear) degree of correlation with human similarity judgments, using the standard Spearman $\rho$ measure.

We used two different instantiations of a vector space model: the first is a 300-dimensional model trained on the Google News corpus, taken from the popular and widely used \texttt{word2vec} package\footnote{\url{https://code.google.com/p/word2vec/}}, which is based on the Skipgram model with negative sampling of Mikolov et al. \cite{mikolov2013distributed}. This model is known to lead to high-quality dense vector embeddings of words. The second space we used is a custom trained 2000-dimensional vector space, trained on the combined UKWaC and Wackypedia corpus, using a context window of 5, and Positive Pointwise Mutual Information as a normalisation scheme on the raw co-occurrence counts. The vectors of this space do not involve any dimensionality reduction techniques, making the vectors relatively sparse compared to those in the \texttt{word2vec} vector space.

For the original dataset, we compare a non-compositional baseline, in which just the vector or matrix for the verb is compared, and additive/multiplicative models, and get the results below:

\begin{table}
{\normalsize
\centering
\begin{tabular}{l@{\hskip 2em}cc}
\toprule
{\bf ML2008} &  				\texttt{word2vec} &  Count Based\\
\midrule
Verb Only Vector &			0.274 & 0.078 \\
Verb Only Tensor &                   0.060 & 0.108 \\
\midrule
Additive         &                          \textbf{0.278} &  0.081 \\
Multiplicative   &                        0.229 & \textbf{0.177} \\
\bottomrule
\end{tabular}
\caption{Spearman $\rho$ correlation scores on the \textbf{ML2008} dataset.}
}
\end{table}

These results are higher than found in the literature: the original evaluation of Mitchell \& Lapata \cite{mitchell2008vector} achieved a highest correlation score of $0.19$, and the regression model of Grefenstette et al. \cite{grefenstette2013multi} achieves a top correlation score of $0.23$. These scores are surpassed already by the non-compositional baseline on the \texttt{word2vec} space here.
Although the highest scores are indeed obtained using a compositional model, note that the correlation for the \texttt{word2vec} model doesn't increase substantially. In the count based space we do see a bump in the correlation when using a compositional model, but here the baseline correlation isn't that high to start with.
The situation is better for the extended dataset. There, we compare the same four models against four combined models, which combine and additive with a multiplicative model, after resolving the ellipsis. The results are in the table below:

\begin{table}[h]
{\normalsize
\centering
\begin{tabular}{l@{\hskip 2em}cc}
\toprule
{\bf MLELLDIS} &  \texttt{word2vec} &  Count Based \\
\midrule
Verb Only Vector &                              0.274 &             0.078 \\
Verb Only Tensor &                              0.060 &             0.108 \\
\midrule
Additive         &                              0.292 &             0.040 \\
Multiplicative   &                              0.068 &             0.206 \\
\midrule
Multiplicative $\odot$         &                              0.213 &             \textbf{0.391} \\
Multiplicative $+$          &                              0.294 &             0.179 \\
Additive $\odot$          &                              0.229 &             0.172 \\
Additive $+$           &                              \textbf{0.298} &             0.078 \\
\bottomrule
\end{tabular}
\caption{Spearman $\rho$ correlation scores on the extended \textbf{ML2008} dataset.}
}
\end{table}

Our first observation is that the naive additive and multiplicative models already do better than the non-compositional baseline, save for the additive model on the count based space. Secondly, even better results are obtained by applying a non-linear compositional model, i.e. a model that actually resolves the ellipsis and copies the representation of the verb. For the case of the \texttt{word2vec} space the best performing model is the fully additive model that adds together all the vectors to give the result $\vek{subj} + \vek{verb} + \vek{subj^*} + \vek{verb}$. For the count based space, it is the exact opposite: the fully multiplicative model achieves the best overall score of $0.391$ with the representation $\vek{subj} \odot \vek{verb} \odot \vek{subj^*} \odot \vek{verb}$.

That the \texttt{word2vec} vectors work well under addition but not under multiplication, whereas the count based vectors work well under multiplication but not under addition, we attribute to the difference in sparsity of the representations: since \texttt{word2vec} vectors are very dense representations, multiplying them will not have a very strong effect on the resulting representation, whereas adding them will have a greater net effect on the final result. In contrast, multiplying two sparse vectors will eliminate a lot of information, since the entries that are zero in one of the vectors leads to a zero entry in the final vector. In other words, the final representation will be incredible specific, allowing for better disambiguation. Addition on sparse vectors however, will simply generate vectors that are very unspecific and are thus not very helpful for disambiguation.

Overall, we see that the presented results are in favour of non-linear compositional models, showing the importance of ellipsis resolution for distributional sentence representations.

\section{Conclusion, Further Work}
\label{sec:conclusion}

In this paper we incorporated a proper notion of copying into a compositional distributional model of meaning to deal with VP ellipsis with anaphora. By decomposing the DisCoCat architecture into a two step interpretation process, we were able to combine the flexibility of the Cartesian structure of the non-linear simply typed lambda calculus, with a vector based representation of word meaning. We presented a vector-based analysis of VP ellipsis with anaphora  and showed how the elliptical phrases get assigned the same meaning as their resolved variants. We also carried out a large scale similarity experiment, showing that verb disambiguation becomes easier after ellipsis resolution. 

By giving up a direct categorical translation from a typelogical grammar to vector spaces, we gain the expressiveness of the lambda calculus, which allows one to interpret the grammatical derivations  in various different concrete compositional models of meaning. We showed that   previous DisCoCat work on resolving ellipsis using coordination and Frobenius algebras \cite{kartsaklisverb} can only be obtained in an ad hoc fashion. For future work we intend to compare the two approaches from an experimental point of view. 

A second challenge that we would like to address in the future involves dealing with derivational ambiguities  in a vectorial setting. These ambiguities were exemplified in this paper by the strict and sloppy readings of elliptical phrases involving anaphora, and cascaded ellipsis. In order to experiment with the vectorial models of these cases, an appropriate task should be defined and experiments should determine which distributional reading can be empirically validated.

Finally,  in previous  work  \cite{wijholdsSadr2017},  we showed how to  resolve ellipsis in a modal Lambek Calculus  which has a controlled form of contraction for formulae marked with the modality. Our work is very similar to an earlier  proposal of J{\"a}ger presented in \cite{jager1998multi}.   The controlled contraction rule that we use is as follows
\[
\infer[]{C(f): A \to B}{f: \Diamond A \otimes A \to B}
\]
The $\Diamond$ modality has a few other rules for controlled associativity and movement. The semantics of this rule is, however, simply defined as $C := \lambda fx. f \langle x, x \rangle$. Trying to find a vector operation (either in a linear setting using a biproduct  operation or by moving to a non-linear setting)  and obtaining a direct categorical semantics is work in progress. The challenge is  that the   interpretations of the similar $!$ modality of Linear Logic, e.g. in  a linguistic setting by Morrill in \cite{Morrill2017} or  in a computational setting by Abramsky in  \cite{ABRAMSKY19933})  would not work in a vector space setting. The  Frobenius algebraic copying operation, with which we  worked  in \cite{wijholdsSadr2017}, is one of the few options available, and we have shown that it does not work when it comes to distinguishing the sloppy versus strict reading of the ambiguous elliptical cases. 


\bibliographystyle{plain}      
\newcommand{\references}{jolli-bibdata}
\bibliography{\references}   


%

\end{document}